\documentclass[10pt,twocolumn,letterpaper]{article}

\usepackage{cvpr}              

\usepackage{graphicx}
\usepackage{amsmath}
\usepackage{amssymb}
\usepackage{booktabs}

\usepackage{algorithm,algpseudocode}
\usepackage{makecell}

\usepackage{arydshln}
\newcommand\norm[1]{\left\lVert#1\right\rVert}

\DeclareMathOperator*{\argmin}{arg\,min}
\algdef{SE}[SUBALG]{Indent}{EndIndent}{}{\algorithmicend\ }%
\algtext*{Indent}
\algtext*{EndIndent}

\usepackage[pagebackref,breaklinks,colorlinks]{hyperref}

\usepackage[capitalize]{cleveref}
\crefname{section}{Sec.}{Secs.}
\Crefname{section}{Section}{Sections}
\Crefname{table}{Table}{Tables}
\crefname{table}{Tab.}{Tabs.}

\def\cvprPaperID{2} 
\def\confName{CVPR}
\def\confYear{2022}

\begin{document}

\title{Nerfels: Renderable Neural Codes for Improved Camera Pose Estimation}


\author{Gil Avraham$^{1}$\thanks{Performed while interning at Facebook Reality Labs}, Julian Straub$^{3}$, Tianwei Shen$^{3}$, Tsun-Yi Yang$^{3}$, Hugo Germain$^{2}$, \\ Chris Sweeney$^{3}$, Vasileios Balntas$^{3}$, David Novotny$^{4}$, Daniel DeTone$^{3}$, Richard Newcombe$^{3}$ \\
Monash University$^{1}$, 
École des Ponts$^{2}$,  
Facebook Reality Labs$^{3}$, Facebook AI Research$^{4}$ \\
{\tt\small gil.avraham@monash.edu}, {\tt\small 	hugo.germain@enpc.fr} \\
\{{\tt\small jstraub, tianweishen, tsunyi, sweeneychris, vassileios, dnovotny,  ddetone, newcombe @fb.com}\}
}

\maketitle

\begin{abstract}
This paper presents a framework that combines traditional keypoint-based camera pose optimization with an invertible neural rendering mechanism. Our proposed 3D scene representation, Nerfels, is locally dense yet globally sparse.
As opposed to existing invertible neural rendering systems which overfit a model to the entire scene, we adopt a feature-driven approach for representing scene-agnostic, local 3D patches with renderable codes. By modelling a scene only where local features are detected, our framework effectively generalizes to unseen local regions in the scene via an optimizable code conditioning mechanism in the neural renderer, all while maintaining the low memory footprint of a sparse 3D map representation. Our model can be incorporated to existing state-of-the-art hand-crafted and learned local feature pose estimators, yielding improved performance when evaluating on ScanNet for wide camera baseline scenarios.
\end{abstract}



\section{Introduction}

The choice of map representation used in Visual Simultaneous Localization and Mapping (Visual SLAM), Structure-from-Motion (SfM) and Visual Localization systems is paramount as it effects the accuracy and power consumption of the system \cite{cadena2016past}. Sparse feature-based approaches \cite{nister04, agarwal2010, mur2015} detect feature points in images which are matched and triangulated across multiple images to form 3D maps. These sparse representations are lightweight and can be effectively used in low power systems like robotics and augmented reality. At the heart of sparse representations is a geometric (also known as indirect) reprojection error, used to optimise camera poses.

\begin{figure}[t]
\centering
\includegraphics[width=0.45\textwidth]{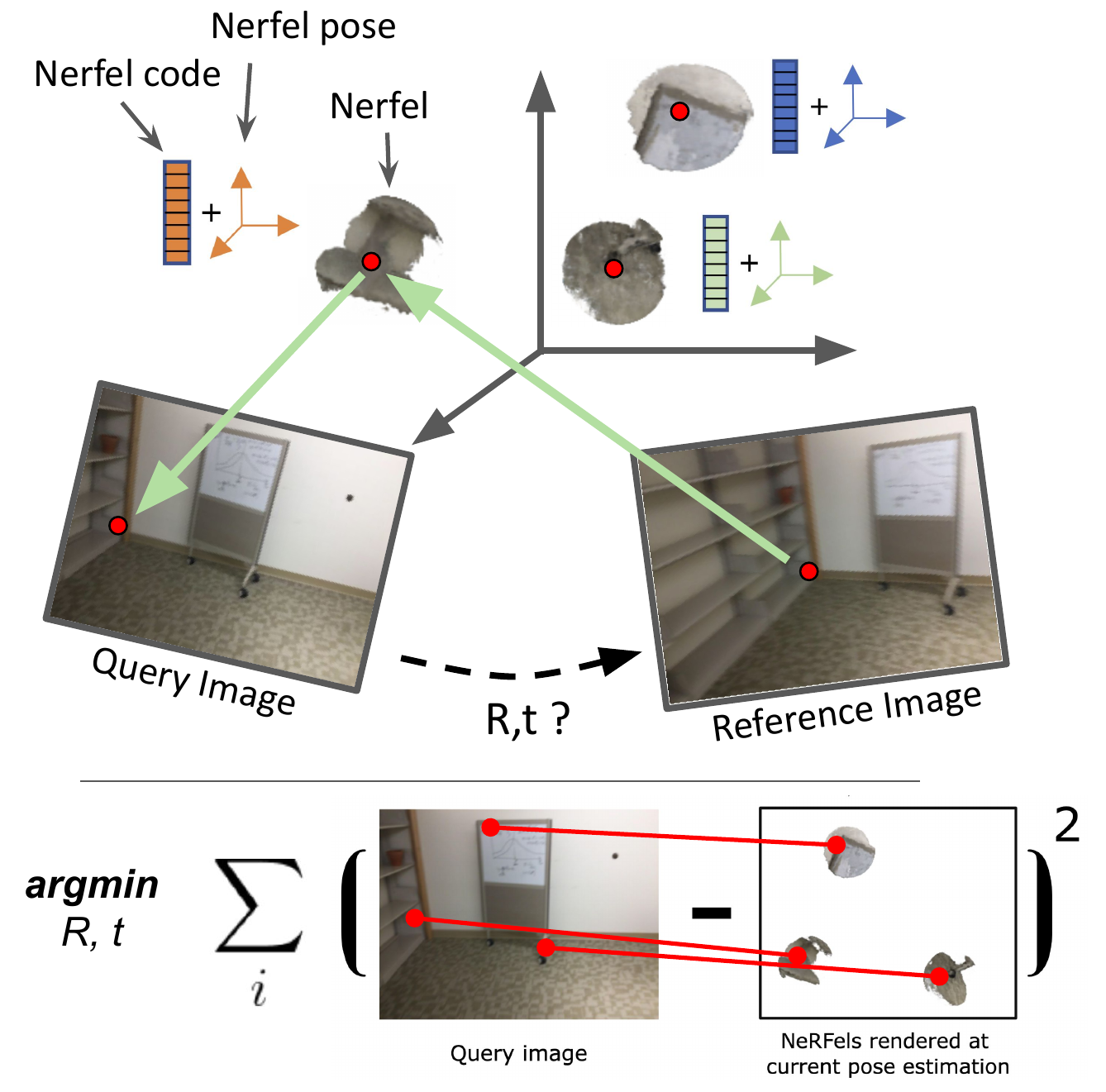}
\caption{\textbf{Camera Pose Estimation with Nerfels}. (top) Nerfels are detected and matched across two views. A local code and pose is initialized for each Nerfel. (bottom) The unknown camera pose is optimised with both a geometric and local photometric loss.}
\label{fig:intro_diagram}
\end{figure}

In contrast, direct methods work with the raw pixel information\cite{newcombe2011dtam}. Dense-direct methods exploit all the information in the image, even from areas where gradients are small; thus, they can outperform feature-based methods in scenes with poor texture, defocus, and motion blur \cite{cadena2016past}. These methods use a photometric alignment objective function similar that used in the well-known Lucas Kanade optical flow algorithm \cite{Lucas81}. Though dense-direct methods can add robustness where sparse feature-based methods struggle, they require good initial solution. This limits their performance in wide-baseline localization where a good initialization is difficult to obtain. Additionally, representing the scene densely results in a larger memory footprint of the map, which can lead to higher power consumption for embedded localization systems\cite{nardi2015introducing}. Thus deciding on the density of the map representation for SLAM and SfM systems can result in trading off various characteristics of accuracy versus power consumption.

We focus our work at the intersection of sparse feature based approaches and dense image alignment approaches by representing the scene in a globally sparse but locally dense manner. We aim to reap the benefits of a lightweight, sparse model that doesn't require good pose initialisation, whilst using the power of generative models that leverage image measurements to constrain the pose estimation. Our approach is inspired by surface elements, also known as Surfels~\cite{pfister2000}, which are point primitives that model scene attributes without explicit connectivity between elements. Surfels are used to efficiently render complex geometric shapes by modelling the scene locally as a small circular plane, and have been used in SLAM systems such as ElasticFusion \cite{whelan15}. In such works, the scene is modelled densely by many Surfels: each individual Surfel models a very small portion of the scene (i.e. one Surfel per pixel observed). An interesting approach might be to use a sparse set of Surfels with a larger extent, and integrate them into pose estimation. However, as the baseline between cameras increases, the local planarity assumption weakens for non-planar scenes, and a simple planar representation does not accurately explain the 3D structure (demonstrated experimentally in \ref{sec:ablation}).

Alternatively, neural rendering offers the promise of novel view rendering for arbitrary 3D scenes. Methods such as NeRF \cite{mildenhall2020nerf} can generate high quality, high resolution images for a limited scale of scenes given enough compute and training views. Code conditioned neural rendering presents an attractive way to train on different scenes and learn generic representations that can work on related but unseen 3D data. Related approaches demonstrate the use of effective code-conditioning to learn re-usable priors of the geometry and appearance \cite{Schwarz2020, remata2021}. Follow up work has shown that one can additionally invert such neural rendering approaches to optimise over the camera pose. However, rendering the scene densely for camera pose estimation can be computationally expensive. One such approach, iNeRF~\cite{lin2020iNeRF}, shows that guiding the rendering to occur near local features can significantly improve run-time. This begs the question of whether it is necessary to keep a dense representation of the scene for pose estimation with neural renderers, or to use a more sparse representation with a lower memory footprint.

To this end, we propose \textit{Nerfels}, a hybrid sparse and dense 3D representation designed for pose estimation that is inspired by neural rendering approaches like NeRF, and locally dense point primitives such as Surfels. Nerfels use a code conditioned neural rendering network to represent the local 3D sphere around a sparse 3D keypoint, resulting in a \textit{locally dense yet globally sparse 3D scene representation}. The use of code conditioning allows each Nerfel to share memory and 3D priors though a single neural network. Each Nerfel has a local pose in the global coordinate frame, which can be explicitly optimized to result in the best rendering across different views. Nerfels are capable of rendering RGB patches around each 3D keypoint into arbitrary views of the 3D scene, which can be used as an additional constraint to a standard sparse 3D point cloud SLAM or SfM pipeline. 
We believe that the Nerfels representation is an effective compromise between sparse and dense representations for embedded SLAM and SfM applications that require high accuracy pose estimation with a small memory footprint for their 3D representation.

\textbf{Contributions}. Our contributions are twofold: (1) We present Nerfels, a novel representation of local shape and appearance of 3D sparse maps that enhances keypoints to be locally renderable; (2) An end-to-end camera pose estimation system using Nerfels through joint optimisation of reprojection error + photometric error, resulting in improvement of wide baseline pose estimation for both hand-crafted and learned local features.




\section{Related Work}

\textbf{Sparse SLAM and SfM} There is a large body of work that tackle camera pose by simultaneously estimating a sparse scene structure from keypoints detected in 2D images. Well-known SfM and SLAM works \cite{klein2007parallel, davison2007, agarwal2011, mur2015} rely on triangulated 3D interest points to build a sparse structure of the scene using handcrafted local features \cite{harris1988, lowe2004distinctive}. 
Neural networks have been used to learn local features \cite{detone2018superpoint, Dusmanu2019CVPR, sarlin2020superglue} that are more robust to challenging scene changes such as lighting and viewpoint change. 
Our work is heavily based on such pose estimation approaches. In challenging pose estimation scenarios however, the accuracy of such approaches can degrade when few matches or inliers exist between the map and the image to be localized. We argue that by leveraging more local image information around the matched keypoints, we can improve pose accuracy.

\textbf{Semi-Dense and Dense SLAM} Early image registration work \cite{Lucas81} used image gradients to align multiple images. More recently, SLAM approaches use direct image alignment \cite{newcombe2011dtam, engel2014lsd, engel2018dso} to jointly estimate a dense or semi-dense scene representation in a real-time pose estimation framework. Such approaches \cite{whelan15, henry2012} incrementally build a map from multiple observations that is able to render physically predictive representations of the scene. More recently, dense neural pose estimation approaches learn a lower dimensional code~\cite{bloesch2018}, basis function~\cite{tang2018ba}, or cost volumes~\cite{zhou2018deeptam} for the depth of each keyframe and optimise the depth jointly with the camera poses.

\textbf{Neural Scene Representation} Structure-less approaches \cite{Kendall2015PoseNetAC, balntas2018relocnet, wang18deepvo} use a neural network to directly estimate camera poses without modelling a map. The localization of cameras is performed through a directly learned mapping from images to poses. More recent works \cite{henriques2018, avraham2018} similarly do not model 3D geometry but instead model a map through neural embeddings. Recent advances in implicit  representation learning  have  demonstrated  the  power of coordinate-based multilayer perceptrons (MLPs) to map known world coordinates to signed distance fields \cite{park2019, jiang2020}, occupancy grids \cite{Mescheder2019OccupancyNL} or RGB  values \cite{mildenhall2020nerf, wang2021ibrnet}. Such systems are capable of modelling physically predictive representations of 3D scenes. In one such work, NeRF, the pose of each camera is computed offline using COLMAP \cite{schoenberger2016sfm}. Follow up works \cite{lin2020iNeRF, lin2021barf, wang2021nerfmm} relax this constraint by optimising both the implicit function scene structure and the camera pose. These works represent the full scene with a single implicit function. The work of iMAP \cite{sucar2021} builds a real-time RGB-D slam system using an implicit neural renderer. We extend this work by applying a similar invertible NeRF framework to multiple smaller NeRF fields simultaneously. Rather than train independent weights for each NeRF field, we use a code conditioning mechanism in an auto-decoder framework, similar to works such as DeepSDF and others \cite{park2019, sucar2021, Schwarz2020, remata2021}. This helps to minimise the memory cost of the neural scene representation.

\section{Nerfels}
\label{sec:nerfels}
\begin{figure}[t]
\centering
\includegraphics[width=0.35\textwidth]{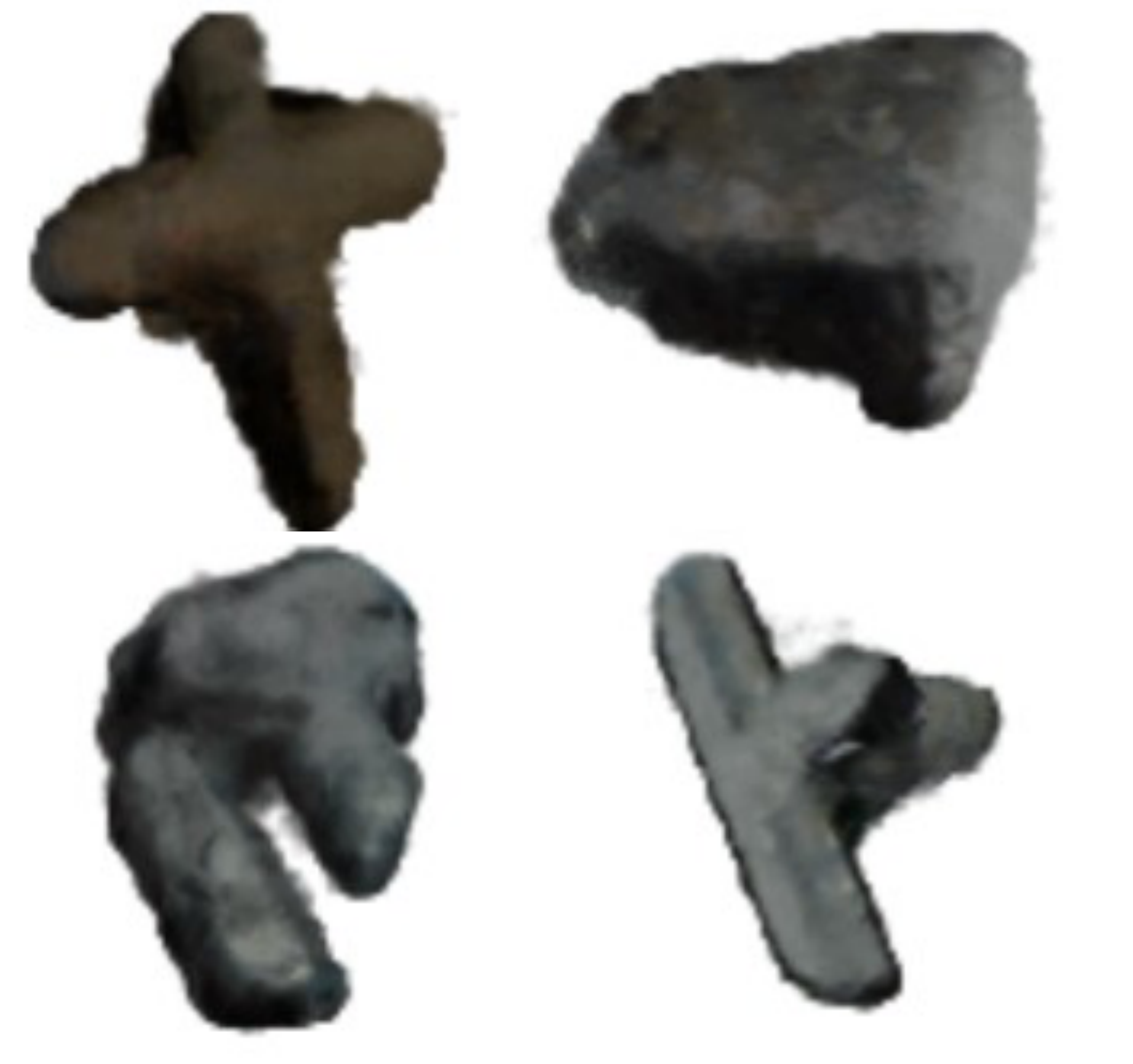}
\caption{\textbf{Examples of Learned Nerfels}. 3D renderings of some example Nerfels are depicted.}
\label{fig:mining_shapes}
\vspace{-0.5em}
\end{figure}

We define a Nerfel as latent code $c \in \mathbb{R}^{d_c}$, to be a compact representation of a continuous radiance field contained in a 3D sphere with radius $r_{s}$. 
In Figure \ref{fig:mining_shapes}, we show an example of $4$ Nerfels mind from a synthetic scene. Obtaining a projection of an individual Nerfels' radiance field to an image canvas, $\bar{I} \in \mathbb{R}^{h \times w \times 3}$, is obtained using a mapping function $\mathcal{D}_{NR}$ which we call neural rendering decoder and a given pose $P_{N} \in \mathbb{SE}(3)$ that the sphere is viewed from: $\bar{I} = \mathcal{D}_{NR}(P_{N}, c)$. In addition, we define $M_{R}$ to be the set of image coordinates where the integrated rays over image $\bar{I}$ have an alpha $\alpha > 0$. The  orientation of a Nerfel is defined by its own canonical coordinate system and isn't bound to any particular global coordinate frame. This coordinate system is chosen so the Nerfels' center of mass is located at the origin, and the orientation is derived by taking the Nerfels' average normal vector and arbitrarily choosing an $up = [0,1,0]^{T}$ vector. \textit{Given single 3D point at the Nerfels'  origin, the Nerfels' derived coordinate system, the camera poses in which the Nerfel is observed and the camera intrinsics, we can compute a set of canonical poses for a single Nerfel.}
For illustration, the shelf looking Nerfel (top left Nerfel) in Figure \ref{fig:mining_shapes} was extracted from a specific shelf of a synthetic frame after aligning the Nerfels' pose to the shelf in the image. However, by changing that Nerfels' pose it can also align to any similar looking shelf which share geometry and appearance.
And thus, a Nerfel can act as a reusable component for describing parts of a scene; whilst one can think of a scene being sparsely described by a collection of Nerfels with a specific pose per Nerfel which aligns each one to that scene.

For the case of rendering a single Nerfel code, the decoder $\mathcal{D}_{NR}$ follows the rendering procedure outlined in ~\cite{mildenhall2020nerf}. In this procedure, a neural-network implicit function parameterised by $\Theta$ maps the viewing pose $P_{N}$ together with a 3D coordinate location $(x,y,z)$ to an RGB value $\hat{r} \in \mathbb{R}^{3}$ and density $\sigma \in \mathbb{R}^{+}$ and is followed by a differentiable volumetric function that outputs the integrated RGB value along the viewing ray from the image canvas. In addition to the viewing direction and 3D location, our decoder also takes a code $c$ which is concatenated to the location positional encoding $\mathcal{L}_{x}$, where the viewing positional encodings are defined by $\mathcal{L}_{\phi}$ \cite{mildenhall2020nerf, tancik2020fourier}. To ease notation we describe the process of rendering a single Nerfel code in a given pose by $\mathcal{D}_{NR}(P_{N}, c)$. We also define the operation of rendering multiple Nerfels as: $\bar{I} = \mathcal{D}_{NR}(\textbf{P}_{N}, \textbf{c})$ where each Nerfel $c_{i} \in \textbf{c}$ rendering provides an image coordinate set $M_{R,i}$; and the union of these sets is: $\boldsymbol{M}_{R} = \bigcup M_{R,i}$. To resolve overlapping Nerfels when obtaining the RGB value in $I$, one can compute the average RGB value of overlapping Nerfels, take a weighted average of Nerfels' RGB and alpha projections or as done in this work arbitrarily select some Nerfels to proceed others (details in Section \ref{sec:experiments}).
\begin{figure}[t]
\centering
\includegraphics[width=0.45\textwidth]{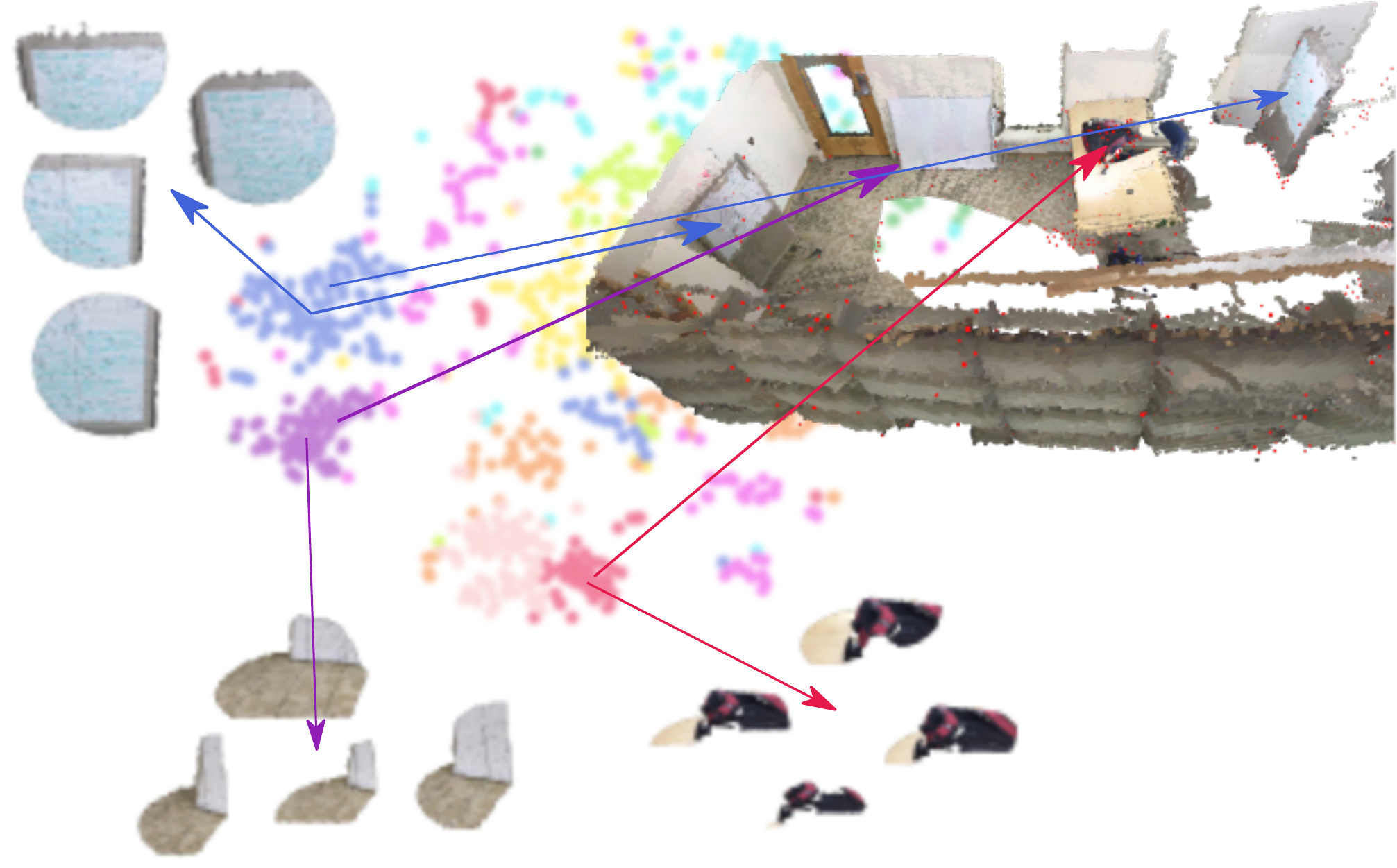}
\caption{\textbf{Visualising Mined Nerfel Codes}. A t-SNE plot of the learned Nerfel codes is visualized with a ScanNet scene. Visually similar parts of the scene learn similar codes.}
\label{fig:tsne_diagram}
\end{figure}
\section{Mining Nerfels}
\vspace{-0.8em}
\label{sec:neural_shape_codes}

\begin{figure*}[t]
\centering
\includegraphics[width=0.85\textwidth]{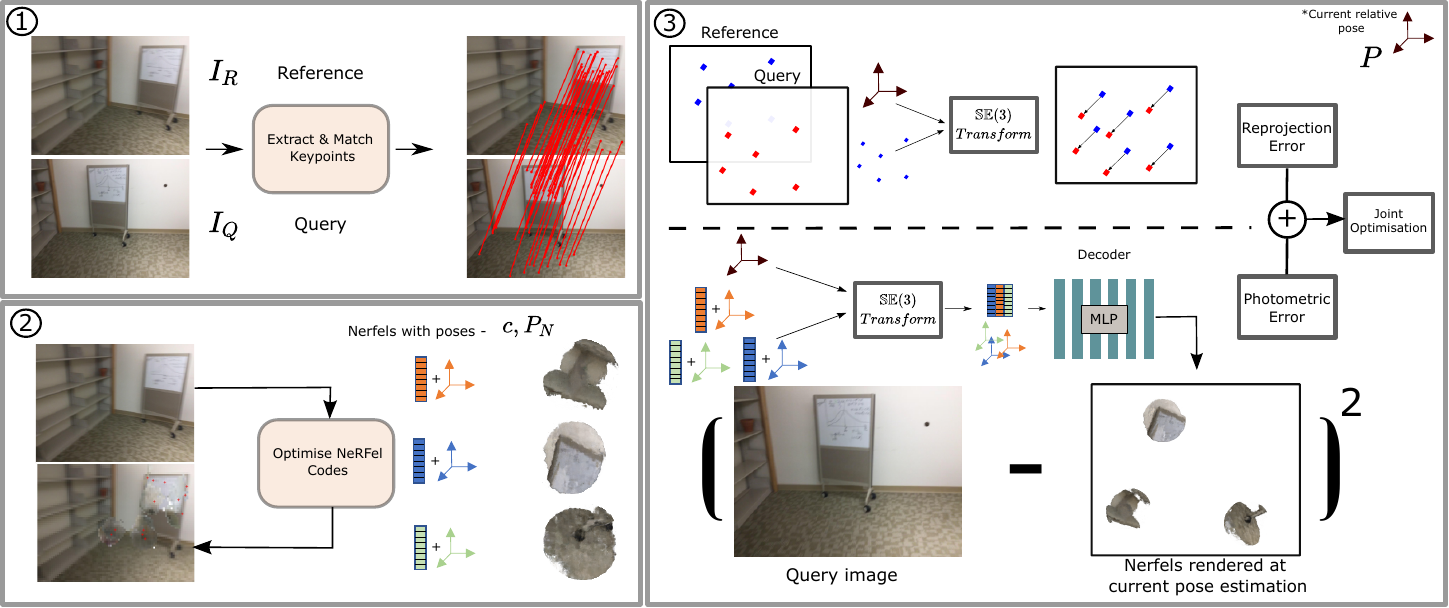}
\caption{\textbf{System Overview of Camera Pose Optimisation with Nerfels}. (1) Keypoints are extracted and matched across two views. (2) the Nerfel codes and poses are initialised. (3) A joint pnp + photometric optimization is performed to get the resulting relative camera pose.}
\label{fig:system_overview}
\end{figure*}

We detail the process of mining Nerfel examples from a collection of scenes and how these are used to train a neural rendering decoder and a set of Nerfel codes. Specifically, we denote a set $N_{c }$ of Nerfel codes as $\mathcal{C} \in \mathbb{R}^{N_{c}\times d_{c}}$ and the decoder as $\mathcal{D}_{NR}(P_{N},c)$. We note two important requirements on our set of codes and the coupled decoder. The first, every code $c$ should encode the associated 3D Nerfel shape, meaning the decoder should be able to render a Nerfel for any given pose, or conversely recover the code (Section \ref{ssec:inerf}) from any provided realisation of the code in an image. The latter, the given set of codes $\mathcal{C}$ should encompass a broad variety of general 3D shapes that are detected with high probability by a given keypoint detector. 

\textbf{Nerfel Examples} For mining a set of Nerfel examples we assume to have access to a collection of different scenes, where each scene has a set of RGB-D frames, $I_{D} \in \mathbb{R}^{h\times w\times 4}$, along with their ground-truth poses $P_{gt} \in \mathbb{SE}(3)$. In addition, we assume to have the intrinsic matrix $K \in \mathbb{R}^{3\times 3}$ which is unique for every scene. The process of collecting a set of Nerfel codes from a given scene is performed by first extracting keypoints in every image within the scene and running a process of retaining the keypoints that can be matched and triangulated between the frames. An example of a method that can extract such matches over a set of frames is COLMAP\cite{schoenberger2016sfm}. These matches can also be obtained by using the ground-truth poses and depth, as done in this work. The result of this process is a set of 3D keypoints $\mathcal{K} \in \mathbb{R}^{N_{kp}\times 3}$ with an additional dictionary of the indices of each keypoint's frame. This representation allows us to sub-select a number of keypoints which appear 
in more than $t_{f}$ frames and are viewed from a wide variety of angles (selection of $t_f$ is discussed in Section \ref{sec:experiments}).
By knowing the 3D location of the remaining keypoints, we crop 2D regions around the keypoint within its observed frames and obtain a collection of patches for each Nerfel in a sequence. For each mined Nerfel, we store a set of image crops $I_{s} \in \mathbb{R}^{h_c \times w_c \times 3}$, where each crop has fixed dimensions $h_c \times w_c$. For each Nerfel with a set of image crops, poses $\mathcal{P}_{N} \in \mathbb{SE}(3)$ are also extracted.
Here, poses $\mathcal{P}_{N}$ are set to be in a canonical coordinate frame so that a Nerfel becomes scene-agnostic.

\textbf{Neural Rendering Decoder} For constructing the neural rendering decoder we make use of neural implicit functions and a volumetric rendering technique discussed in Section \ref{sec:nerfels}. The implicit function is implemented by a MLP and is parameterised by $\Theta$ while the Nerfel codes are initialised as $c_{i} \sim \mathcal{N}(0, \frac{1}{d_{c}})$. The implicit function parameters and the Nerfel codes are optimised as follows:
\begin{equation}
\hat{\Theta}, \hat{\mathcal{C}} = \argmin_{\Theta, \mathcal{C}}\mathcal{L}_{D}(\Theta, \mathcal{C}),  
\quad\mathcal{C} = \{c \in \mathbb{R}^{d^c} | \norm{c}_{2}^{2} = 1\},
\label{eq:shape_NERF_optimisation}
\end{equation}
where $\mathcal{L}_{D}$ is the NeRF\cite{mildenhall2020nerf} objective function modified with a code-condition mechanism described in Section \ref{sec:nerfels}. The RGB observations that are used in the optimisation procedure are the mined Nerfel crops $\mathcal{I}_{s}$ together with the poses $\mathcal{P}_{N}$.
We note that in the optimisation of Equation \ref{eq:shape_NERF_optimisation} a coarse to fine strategy is used; we refer the reader to \cite{mildenhall2020nerf} for further details. While we follow the MLP architecture from \cite{mildenhall2020nerf}, the input to the implicit function is of dimension $\mathcal{L}_{x} + d_{c}$ where positional encodings and the Nerfel code are concatenated before being fed to the implicit function. We note in Equation \ref{eq:shape_NERF_optimisation} the joint optimisation of the implicit function parameters $\Theta$ and the set of codes $\mathcal{C}$. 
By allowing the Nerfel codes to be optimised jointly, similar Nerfels collected from the objects with similarities or partial spatial overlap are pulled together (see Figure \ref{fig:tsne_diagram}).
In the procedure of mining Nerfels, by adjusting the threshold $t_{f}$, we can control the minimal number of image patches per Nerfel. 

\section{Nerfels for Pose Estimation}
The following sections detail how Nerfels are leveraged for camera pose estimation. Figure \ref{fig:system_overview} provides a high-level diagram of our Nerfel-based camera estimation pipeline.

\subsection{Optimising Nerfel Codes}
\label{ssec:inerf}
For an incoming reference image with detected keypoints, Nerfel codes are also extracted and associated with the extracted keypoints (In Figure \ref{fig:system_overview} this is illustrated in 2). In section \ref{sec:neural_shape_codes} we constructed a set of codes $\mathcal{C}$ and a decoder $\mathcal{D}_{NR}$ in a process that is strongly coupled with the keypoints used to extract the Nerfels' information for optimising the decoder. 
Following the work of \cite{yen2020inerf} we perform an inverted optimisation using the decoder $\mathcal{D}_{NR}$ for recovering the Nerfel code $c$ and its canonical pose $P_{N}$ from an RGB measurement of a matched keypoint $\boldsymbol{k}$ in the reference image. This optimisation objective is formulated as:
\begin{equation}
\begin{split}
\hat{P}_{N}, \hat{c} &= \argmin_{P_{N}, c} \mathcal{L}_{inv}(P_{N}, c), \\ 
\mathcal{L}_{inv}(P_{N}, c) &= \sum_{i,j \in \mathcal{I}} \norm{\mathcal{D}_{NR}(P_{N},c)[i,j] - I_{k}[i,j]}^{2},
\end{split}
\label{eq:iNERF_optimisation}
\end{equation}
where $I_{k}$ is an image patch taken around a keypoint $\boldsymbol{k}$ from a reference frame. We note that the recovered code $\hat{c} \notin \mathcal{C}$ may not be a member of the set of training codes $\mathcal{C}$.

For optimising Equation \ref{eq:iNERF_optimisation} we refer the reader to the pose parameterisation discussed in \cite{yen2020inerf} where the pose $P_{N} = e^{[\phi]\theta}P_{0}$ is parameterised in exponential coordinates:  to ensure the pose is a valid $\mathbb{SE}(3)$ member. 
In practice, to improve convergence,
we optimise Equation \ref{eq:iNERF_optimisation} a fixed number of times by sampling different initial poses $P_0$ on a sphere. The pose parameters are initialized by drawing $\phi \sim \mathcal{N}(0, 10e^{-6})$, and the Nerfel code $c \sim \mathcal{N}(0, \frac{1}{d_{c}})$.

\subsection{Joint PnP and Photometric Pose Optimisation}
\label{ssec:joint_pnp_photo}
Given reference and query frames $I_{R}$ and $I_{Q}$ respectively, we extract and match $N_{m}$ keypoints $\boldsymbol{k}_{R}$ and $\boldsymbol{k}_{Q}$.
When formulating the PnP problem \cite{lepetit2009epnp}, we assume a model-based approach where the keypoints in the reference frame represent a map with sparse 3D reconstruction; this provides us with depth values $\boldsymbol{z}_{R}$ associated with keypoints $\boldsymbol{k}_{R}$. 
Given the matched keypoints $\boldsymbol{k_{R}}, \boldsymbol{k_{Q}}$, and $\boldsymbol{k}_{R}$'s Nerfel codes $\boldsymbol{c}$, we formulate the following optimization objective:
\begin{equation}
\hat{P} = \argmin_{P} \mathcal{L}_{PnP+Photo}(P, \boldsymbol{k}_{R}, \boldsymbol{k}_{Q}, (\boldsymbol{P}_{N}, \boldsymbol{c})),
\label{eq:Joint_pnp_photometric}
\end{equation}
where $\mathcal{L}_{PnP+Photo}$ is:
\begin{equation}
\begin{split}
\mathcal{L}_{PnP+Photo} = \sum_{i=1}^{N_{m}}\norm{\pi(P\pi^{-1}(k_{R,i},z_{R,i})) - k_{Q,i}}^{2} + \\
\lambda_{Photo}\frac{1}{|\boldsymbol{M}_{R}|} \sum_{i,j \in \boldsymbol{M}_{R}}\norm{\mathcal{D}_{NR}(P\textbf{P}_{N},\textbf{c})[i,j] - I_{Q}[i,j]}^{2}.
\end{split}
\label{eq:Joint_pnp_photometric}
\end{equation}
The operators $\pi(\cdot)$ and $\pi^{-1}(\cdot)$ denote the perspective projection and unprojection operators respectively.
The objective of Equation \ref{eq:Joint_pnp_photometric} is to recover the pose $P \in \mathbb{SE}(3)$ between a reference and a query frame (In Figure \ref{fig:system_overview} this is illustrated in 3). The first component in this objective is the familiar re-projection error used in PnP, where the reference frame keypoints are projected to the query frame and the error is computed between the keypoints in image coordinates. The second component is a photometric error balanced with a regularisation term $\lambda_{Photo}$. The Nerfel codes $\textbf{c}$ and their canonical pose $\textbf{P}_{N}$ in the reference frame are jointly rendered onto the query frame using the current pose estimate $P$. 
The decoder $\mathcal{D}_{NR}$ is a neural network with a differentiable volumetric rendering, and hence recovering the pose $P$ from Equation \ref{eq:Joint_pnp_photometric} can be performed using gradient-descent with modern Autograd libraries \cite{paszke2019pytorch}.

\section{Experiments}
\label{sec:experiments}
We evaluate our proposed method on a synthetic dataset using a simplified version of the method. This is followed by a real world dataset Scannet\cite{dai2017scannet} evaluation with the full version of the method. When optimising the decoder, the available training data is RGB-D with ground truth poses. When evaluating, we use only RGB data with sparse depth for the reference image keypoints. For details regarding the network architecture used for the neural renderer decoder, training and evaluation settings refer to the supplementary material.

\textbf{Hyperparameters} For recovering the Nerfel codes when computing Equation \ref{eq:iNERF_optimisation} per Nerfel code, we sample $16$ pose initialisations on a sphere to maximise the probability of detecting the correct Nerfel pose $P_{N}$. For each reference image, we limit the number of Nerfel codes to $6 \ll N_{m}$, due to the expensive nature of storing the decoder gradients for each Nerfel during the joint optimisation performed in Equation \ref{eq:Joint_pnp_photometric}. For sub-selecting Nerfel codes, we use a similar strategy as done in \cite{mur2015} where we take a grid of $8 \times 6$, select the highest scoring keypoint in a grid cell and finally take the top scoring keypoints according to the amount of Nerfel limit. This procedure ensures Nerfels are well spaced out and land on a surface with a high chance of good fidelity when rendering, while avoiding the case where a Nerfel might occlude another Nerfel.
For the real-world dataset, we set $\lambda_{Photo} = 1000$ and $\lambda_{Photo} = 10000$ for the synthetic one.


\textbf{Runtime} We consider our method to be model-based approach meaning we assume the Nerfel pose recovery discussed in Section \ref{ssec:inerf} can be performed offline and we report the run-time of performing the joint optimisation (Section \ref{ssec:joint_pnp_photo}) to be $0.137\frac{iters}{sec}$ (which is directly affected by the number of Nerfels used during the optimisation procedure). Note that due to the least square objective function, run-time can be further optimized using a second order solver such as Levenberg-Marquardt using packages such as Ceres Solver \cite{ceres-solver}. Additionally, because rendering occupies the majority of the compute time for our method, additional NeRF speed-up approaches such as \cite{reiser2021ICCV, garbin2021, hedman2021snerg} should significantly reduce run-time. We leave additional run-time optimisation for future work.

\begin{table*}[h]
\small
\begin{center}
\setlength\tabcolsep{0.1cm}
\scalebox{1.0}
{
\begin{tabular}{c c c c c c c c c c c c}
\specialrule{.2em}{.1em}{.1em}
\thead{Features} & \thead{Matcher} & \thead{Translation \\ error (m)} &\thead{Angle \\ error ($^{\circ}$)} &  \thead{Translation \\ error (m)} &\thead{Angle \\ error ($^{\circ}$)}  & \thead{Translation \\ error (m)} &\thead{Angle \\ error ($^{\circ}$)} \\
\hline
  &   & @0.25m & @0.25m & @0.5m & @0.5m & @1m & @1m \\
\hline
SIFT & NN + Ratio test & 0.0582	& 2.6863 & 0.0687 & 3.2767 & 0.0826 & 5.6018 \\
SIFT + Nerfels (Ours) & NN + Ratio test & \textbf{0.0568} & \textbf{2.1422} & \textbf{0.0628} & \textbf{2.2202} & \textbf{0.0717} & \textbf{2.8391} \\ \hdashline[3pt/3pt]
D2Net & NN + Ratio test & 0.0565 & 1.5363 & 0.0626 & 1.7036 & 0.0786 & 2.4709 \\
D2Net + Nerfels (Ours) & NN + Ratio test & \textbf{0.0506} & \textbf{1.3034} & \textbf{0.0532} & \textbf{1.3737} & \textbf{0.0608} & \textbf{1.4878} \\ \hdashline[3pt/3pt]
SP & Superglue & 0.0489 & 1.4317 & 0.0502 & 1.4669 & 0.0502 & 1.4669 \\
SP + Nerfels (Ours) & Superglue & \textbf{0.0389} & \textbf{1.0655} & \textbf{0.0389} & \textbf{1.0655} & \textbf{0.0389} & \textbf{1.0655} \\ \hdashline[3pt/3pt]
Dense pixels & iNeRF & 0.1487 & 53.9162 & 0.2689 & 75.1078 & 0.4262 & 73.0027 \\
SP + Colour & Superglue & 0.0463 & 1.423 & 0.0535 & 1.6259 & 0.0547 & 1.6489 \\
\hline
\end{tabular}
}

\caption{\textbf{Pose Estimation Results on ScanNet}. Nerfels is combined with various classical and learned detectors + descriptors, and pose estimation results are compared. Te cumulative translation and angle errors are reported at different translation error cut-off points. Nerfels improves pose estimation accuracy.}
\label{tab:scannet_results}
\end{center}
\end{table*}

\textbf{Error Metrics}
The errors we inspect are the translation error in metric scale and rotation error in degrees. For both of these error metrics, we examine the error in different cut-off thresholds, specifically: 0.25m, 0.5m and 1m of translation error. 
\subsection{Synthetic Results}
We use a synthetic dataset to motivate the joint optimisation procedure discussed in Section \ref{ssec:joint_pnp_photo}. To do so, we simplify the objective by replacing the neural rendering decoder $\mathcal{D}_{NR}$ in Equation \ref{eq:Joint_pnp_photometric} with a differentiable renderer that requires ground-truth depth for rendering a simulated Nerfel.
This will bypass the neural rendering component which is required for recovering the Nerfel codes and re-rendering the Nerfel codes at different poses throughout the joint optimisation.  
By using the ground-truth depth we extract a simulated Nerfel, realised by a coloured point cloud, around each key-point from the reference frame. This means the Nerfels are extracted ``on the fly" and there is no need to pre-train a decoder.

\begin{figure}[t]
\centering
\includegraphics[width=0.45\textwidth]{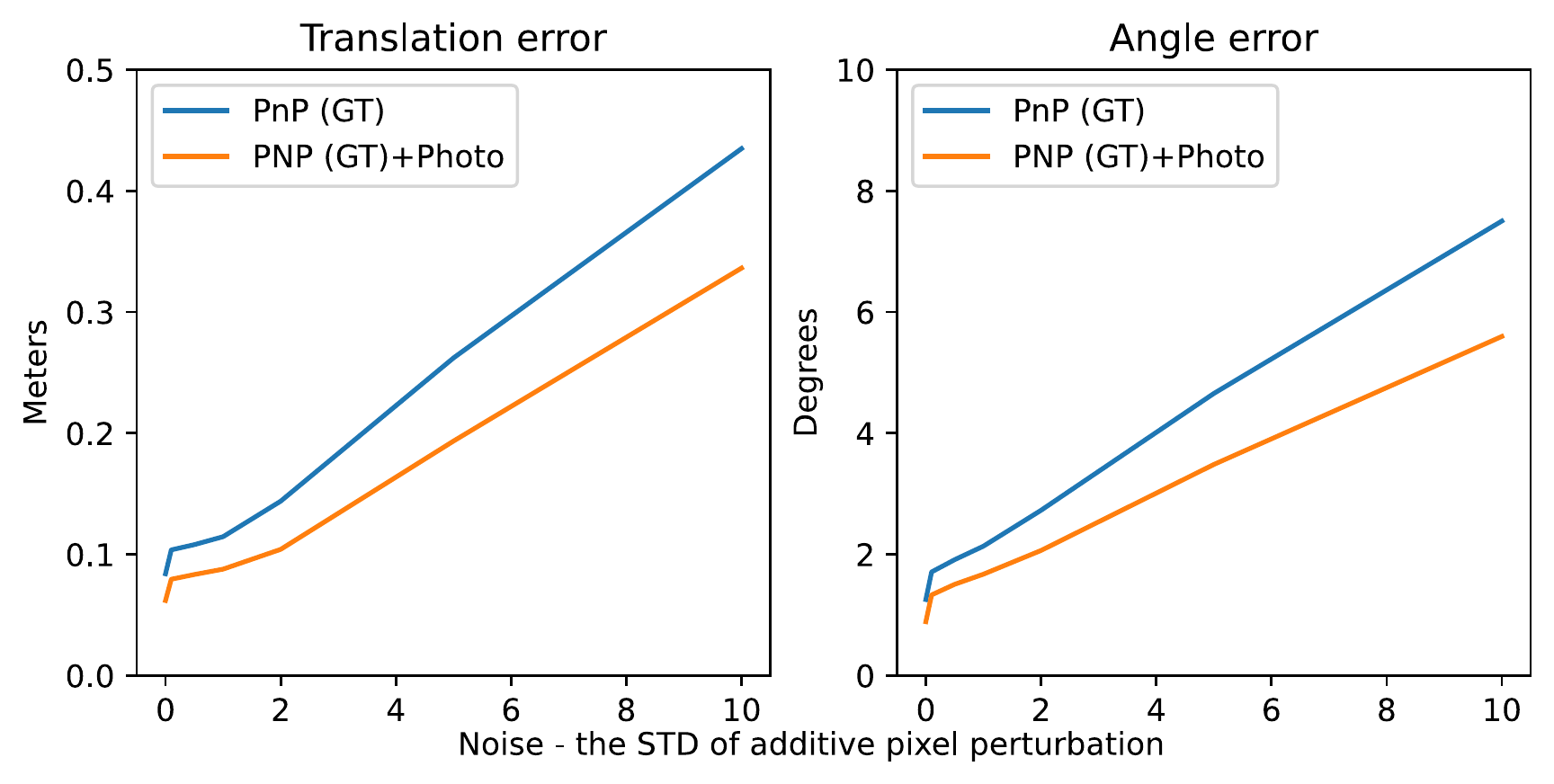}
\caption{\textbf{Joint Geometric and Photometric Ground Truth Analysis}. To simulate the effect of Nerfels on pose estimation, the ground truth depth is used to extract a coloured point-cloud to simulate Nerfel rendering. Translation (left) and rotation (right) errors are plotted as function of additive keypoint noise, simulating poorly localized keypoints.}
\label{fig:all_cases_scenenet_plot}
\end{figure}

The Nerfels are rendered at each iteration using the PyTorch3D Pulsar renderer \cite{ravi2020pytorch3d, lassner2020pulsar} with the current pose estimation. 
For optimising Equation \ref{eq:Joint_pnp_photometric}, we use Adam with a learning rate of $1e^{-2}$, exponential decay of $0.8$ and the number of optimisation iterations performed was $iters=1000$. The plots in Figure \ref{fig:all_cases_scenenet_plot} show the results for this experiment. For this experiment we use SIFT \cite{lowe2004distinctive} and a ground-truth matcher that uses the available ground-truth depth to provide the matches. The purpose of this experiment is to exemplify better robust behaviour of using the extracted Nerfels when the keypoints are not well localised. To do so noise is added to the ground-truth matches' keypoints locations to simulate localization error from real-world matches. Figure \ref{fig:all_cases_scenenet_plot} shows that when noise is added to the keypoints location, adding the photometric term to the optimisation (Equation \ref{eq:Joint_pnp_photometric}) helps recover a more robust solution in case of inaccurate keypoint detection. This simulated experiment helps motivate how Nerfels are useful: \textit{Nerfels provide a boost to pose estimation accuracy that increases with more keypoint detector localization error.}

\subsection{Real-World Results}


\begin{figure*}[t]
\centering
\includegraphics[width=0.85\textwidth]{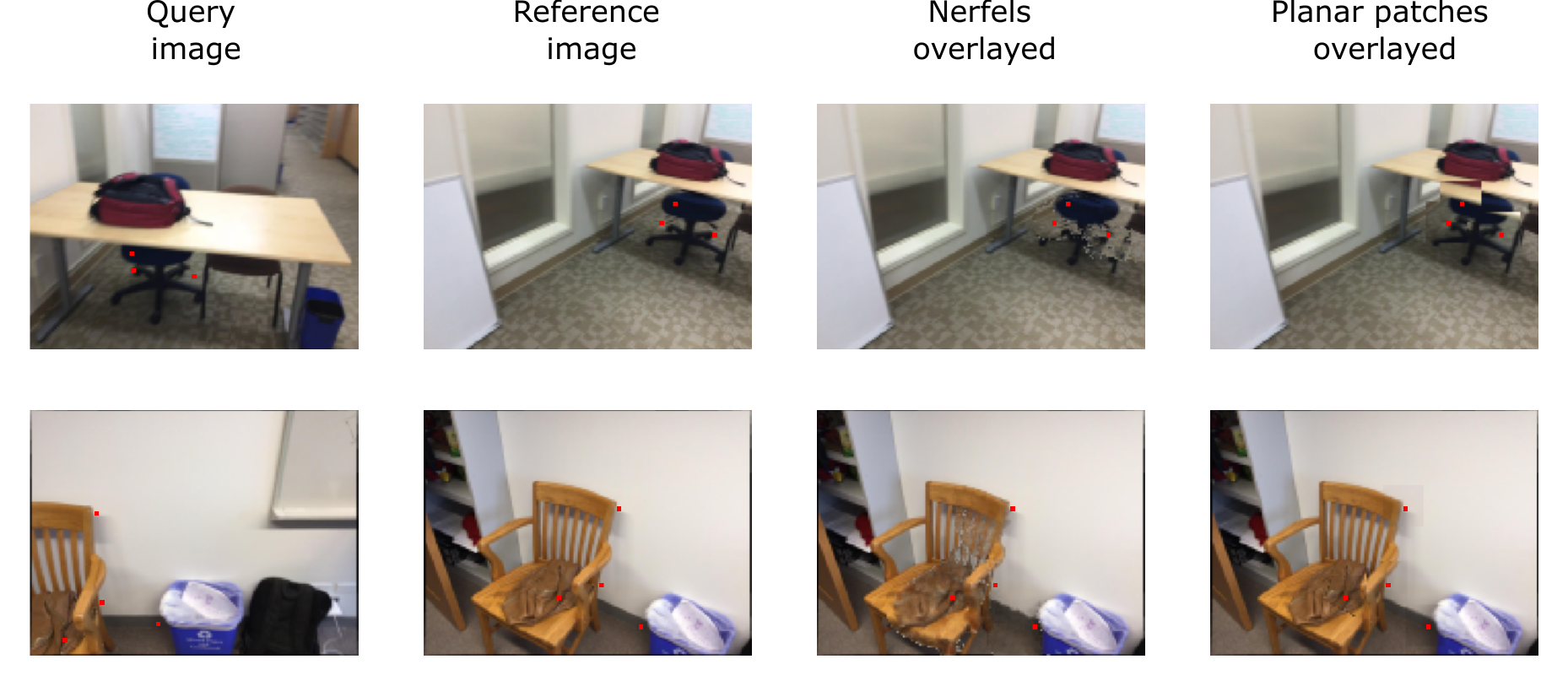}
\caption{\textbf{Qualitative Results of Nerfel Renderings}. Each row shows a different camera pose estimation example. (left) Query image with keypoints. (left-middle) Original reference image with keypoints. (right-middle) Nerfel codes on the reference image using query image codes.  (right) Planar patches on the reference image using query patches. In the top row we point to the distorted planar patch at the edge of the table. In the bottom row, the chair's arm rest can't be correctly aligned using only a planar reprojection while the Nerfels capture the correct 2D projected shape and texture in the keypoints' transformed neighbourhood.}
\label{fig:surfel_fig}
\end{figure*}

We evaluate our method on ScanNet\cite{dai2017scannet} by selecting $8$ scenes and from those $500$ image pairs.

\textbf{Comparison to Sparse Methods} We select three off-the-shelf feature detectors for these experiments: SIFT\cite{lowe2004distinctive}, D2Net\cite{Dusmanu2019CVPR} and SuperPoint\cite{detone2018superpoint} (SP) as a baseline for adding the Nerfel component in the joint optimisation. A separate Nerfels network is trained for each scene as described in Section \ref{sec:neural_shape_codes}, resulting in 8 Nerfel models for each feature detector method, and 24 Nerfel models in total. These feature detectors provide a diverse assessment of classical and learned feature detectors. SIFT detects pixel accurate keypoints though provides few matches whilst learned methods, SuperPoint and D2Net, detect increased amount of keypoints though offer less pixel accurate keypoint with far more matches \cite{dusmanu2020multi}. For feature matching of SIFT and D2Net we use nearest neighbor (NN) + ratio test as described in \cite{lowe2004distinctive} with the ratio test values taken to be 0.7 and 0.9 respectively. For SuperPoint we use a recent state-of-the-art matcher Superglue\cite{sarlin2020superglue} which was developed and evaluated to handle similar camera baseline cases with the same overlap ratio. For evaluating our Nerfel over each of the baselines we use Adam as the optimiser, a learning rate of $1e^{-2}$, exponential decay of $0.8$ and the number of optimisation iterations were $iters=200$. \\
For each sparse matching method, we run two pose estimation experiments: one with and one without Nerfels. Inspecting Table \ref{tab:scannet_results}, we note that adding the Nerfels via the joint optimisation improves results across all baselines. While D2Net baseline provides generally better results over the SIFT baseline, it is considered a less pixel accurate detector. Adding the Nerfels to D2Net provides a better relative performance vs. adding Nerfels to SIFT which in the 0.25m regime provides a modest improvement. Across all results the largest gains are seen when taking the error at a larger cut-off point (largest @$1m$). This improvement aligns well with the empirical results seen in the synthetic data and stems from the Nerfels ability to assist in cases where the error is large.

\textbf{Comparison to iNerf} A vanilla iNeRF\cite{yen2020inerf} network was trained to capture an entire scene using the same network architecture and capacity that is used in our rendering decoder. iNerf as originally presented does not use depth information, but our method does. To make the comparison more fair, we we modify the iNeRF loss by adding a depth loss, similar to \cite{sucar2021}. Accounting for our code conditioning mechanism, for a representative number of roughly $500$ codes at training time, and a 64 dimensional code, our Nerfel model capacity is $3\%$ higher than a Vanilla NeRF. By using a NeRF network to represent an entire scene we can then use iNeRF\cite{yen2020inerf} for pose estimation. In comparison to our runtime, iNeRF runs at a $0.207\frac{iters}{sec}$.

For iNeRF, we generally saw that when less observations are provided in certain regions of a scene, pose estimation fails. This aligns well with how NeRF relies on dense sampling of radiance fields. On the contrary, our Nerfels by construction do not suffer from this issue, as the mining procedure enforces a dense radiance field when selecting keypoints for constructing Nerfel codes. Please refer to the supplementary material for qualitative results of our mined Nerfels and dense scene representation using a vanilla NeRF.\\

\subsection{Ablation studies}
\label{sec:ablation}
\textbf{Studying the joint optimisation} We ran a study where the optimisation was initialised with the PnP result of SuperPoint+Superglue. The reprojection was turned off and only the Photometric component was used. When the initial pose was above 0.5m the optimisation failed in the majority of the cases. In the cases of below 0.25m partial success was noted with results of $Translation \, error = 0.069m$ and $Angle \, error = 3.3231^{\circ}$. 
We conclude that the reprojection error is required to sufficiently provide a geometric constraint on Photometric term in all cases.

\textbf{Na\"ive colour baseline} To study the effect of importance in rendering fidelity we propose a na\"ive approach which samples the keypoint colour for the matching keypoints and performs the joint optimisation while using the colour value in the Photometric component instead of the Nerfels. The results are shown in Table \ref{tab:scannet_results} as ``SP + Colour". The empirical results indicate that when the pose initialisation is below the 0.25m this approach can improvement the vanilla PnP approach. However for pose initialisations above 0.5m performance degrades as the reference and query images might exhibit significant colour changes.

\begin{figure}[t]
\centering
\includegraphics[width=0.4\textwidth]{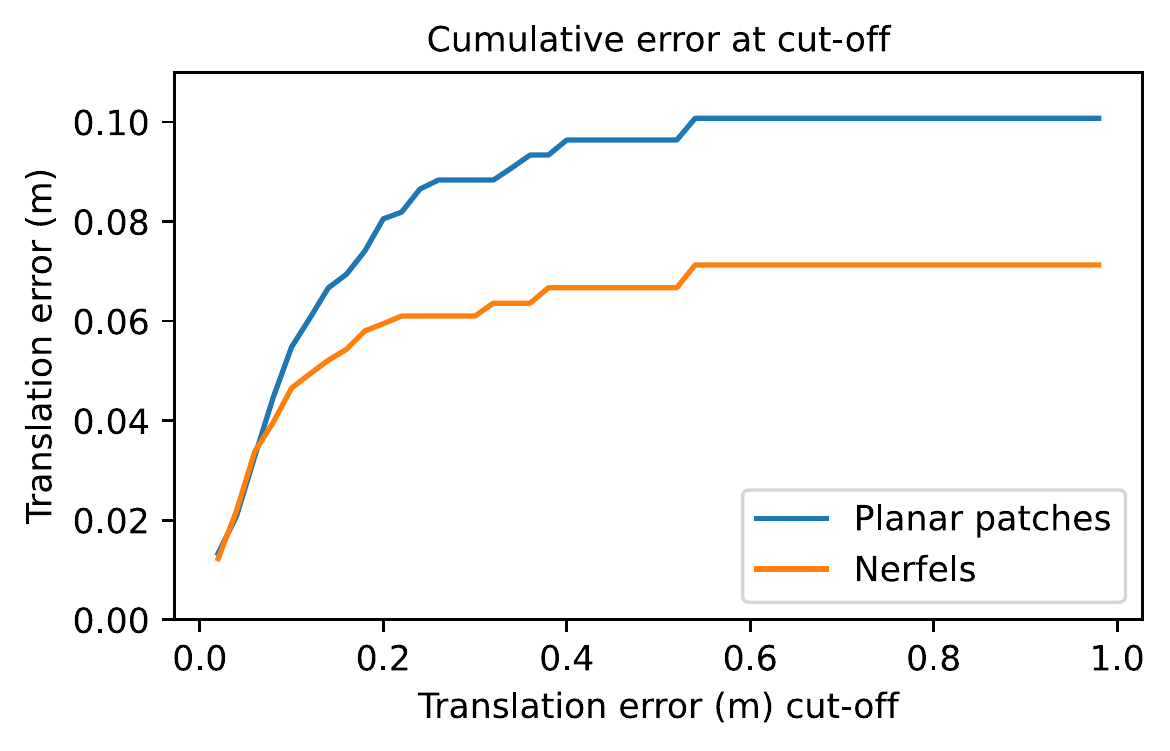}
\caption{\textbf{Evaluating Nerfels vs. planar patches.} In low error regions both perform similarly. As the error grows Nerfel out-performs the planar patches method as it's able to render the rigid changes in viewpoints.}
\label{fig:surfels_exp}
\end{figure}

\textbf{Nerfels vs. Planar patches} The main strength of the Nerfel code is its ability to capture the 3D neighbourhood around a keypoint. We compare how the Nerfel code evaluates against planar patches sampled around a keypoint and re-rendered at the optimsation phase by assuming the patch's normal is orthogonal to the camera. For incorporating the planar patches into the photometric term in Equation \ref{eq:Joint_pnp_photometric} we interpolate the pixel values given the projected patch. To highlight the benefits of the Nerfel code over planar patches, we re-select 100 test cases now with an overlap of $[0.4, 0.6]$. In addition, to faithfully compare planar patches vs. the 3D Nerfel spheres we by-pass optimising the code and codes' pose and assume these were recovered. In Figure \ref{fig:surfels_exp} we see a cumulative error plot where as the cut-off error increases the gap between using Nerfels vs. planar patches widens. This is due to Nerfels' ability to render the rigid behaviour of the keypoints' neighborhood in wider baseline cases as opposed to the planar patches (Qualitative examples shown in Figure \ref{fig:surfel_fig}).  

\section{Limitations}
Our optimisation procedure on average runs at: $0.137\frac{iters}{sec}$. While providing an improvement over iNeRF, our method does not run in real-time when using a standard SGD optimiser (further discussed in Section \ref{sec:experiments}). When optimising Equation \ref{eq:Joint_pnp_photometric} we use a photometric loss which is sensitive to illumination changes. We leave improvements on this matter to future work.
\section{Conclusion}
In this paper, we presented a framework for combining traditional geometric feature matching for pose estimation with local photometric alignment. The photometric alignment is performed using Nerfels, which use an underlying code-conditioned neural rendering mechanism. We verified experimentally that the additional constraints from the local photometric alignment improve pose estimation, especially in wide baseline scenarios. One key characteristic of this formulation is that by using a code-conditioned NeRF, we maintain the advantage of low-memory footprint maps, which is critical for applications like AR and Robotics, while gaining the advantages of a partially generative model. Additionally, relative to full scene neural rendering approaches, we reduce the required expressively of the neural renderer to local parts of the scene, enabling improved generalisation of the approach to unseen parts of the scene. 


\newpage

{\small
\bibliographystyle{ieee_fullname}
\bibliography{egbib}
}

\clearpage
\appendix

\section{Algorithm Outline}
\begin{algorithm}
\caption{A full algorithmic outline of proposed system}
\begin{algorithmic}
  \State $\mathbf{OFFLINE}$
  \State $\{\mathcal{I}_{i}, \mathcal{P}_{i,N}, c_{i}\}_{i=1}^{N_{c}} \leftarrow$ Extract triplets of images poses and codes ($c_{i} \sim \mathcal{N}(0, \frac{1}{d_{c}})$) from a scene
  \State $\mathcal{D}_{NR}(P_{N},c) \leftarrow$ Optimise $\mathcal{L}_{D}$ to recover $\hat{\Theta}, \hat{\mathcal{C}}$
  \\
    \State $\mathbf{ONLINE}$   
    \State RelativePoseEstimator($I_{R}, I_{Q}$):
    \Indent
         \State \# Extract keypoints and features
         \State $\boldsymbol{k}_{R}, \boldsymbol{d}_{R} \leftarrow keypointFeatureExtractor(I_{R})$
         \State $\boldsymbol{k}_{Q}, \boldsymbol{d}_{Q} \leftarrow keypointFeatureExtractor(I_{Q})$
         \State \# Match keypoints and retain the matches
         \State $\boldsymbol{k}_{R}, \boldsymbol{k}_{Q} \leftarrow Matcher(\boldsymbol{k}_{R}, \boldsymbol{d}_{R}, \boldsymbol{k}_{Q}, \boldsymbol{d}_{Q})$
         \State \# Extract reference Nerfels)
         \For{$i=1$ to ${|\boldsymbol{k}_{R}|}$} 
         \State $\hat{P}_{N,i}, \hat{c}_{i} \leftarrow \argmin_{P_{N,i}, c_{i}}\mathcal{L}_{inv}(P_{N,i}, c_{i}, k_{i})$
         \EndFor
         \State \# Solve relative pose
         \State $\hat{P} \leftarrow \argmin_{P}\mathcal{L}_{PnP+Photo}(\boldsymbol{k}_{R}, \boldsymbol{k}_{Q}, (\hat{\boldsymbol{P}}_{N}, \hat{\boldsymbol{c}}))$
         \State \Return $\hat{P}$
    \EndIndent
\end{algorithmic}
\label{alg:full_algorithm}
\end{algorithm}

\section{Cumulative Error Curves}

We provide cumulative error curves corresponding to the feature extractors from Table 1 (SIFT, D2Net, Superpoint). These curves add additional detail to the evaluation performed in the main paper, which selected operating points of 0.25, 0.5, and 1.0 meters along the x-axis. In each of the three cases, the addition of the photometric term into the optimisation provided by Nerfel rendering improves across all thresholds. Additionally, as the difficulty of the pose estimation increases, the gap widens. 


\begin{figure}[t]
\centering
  \begin{tabular}{c}
  \includegraphics[width=0.3\textwidth]{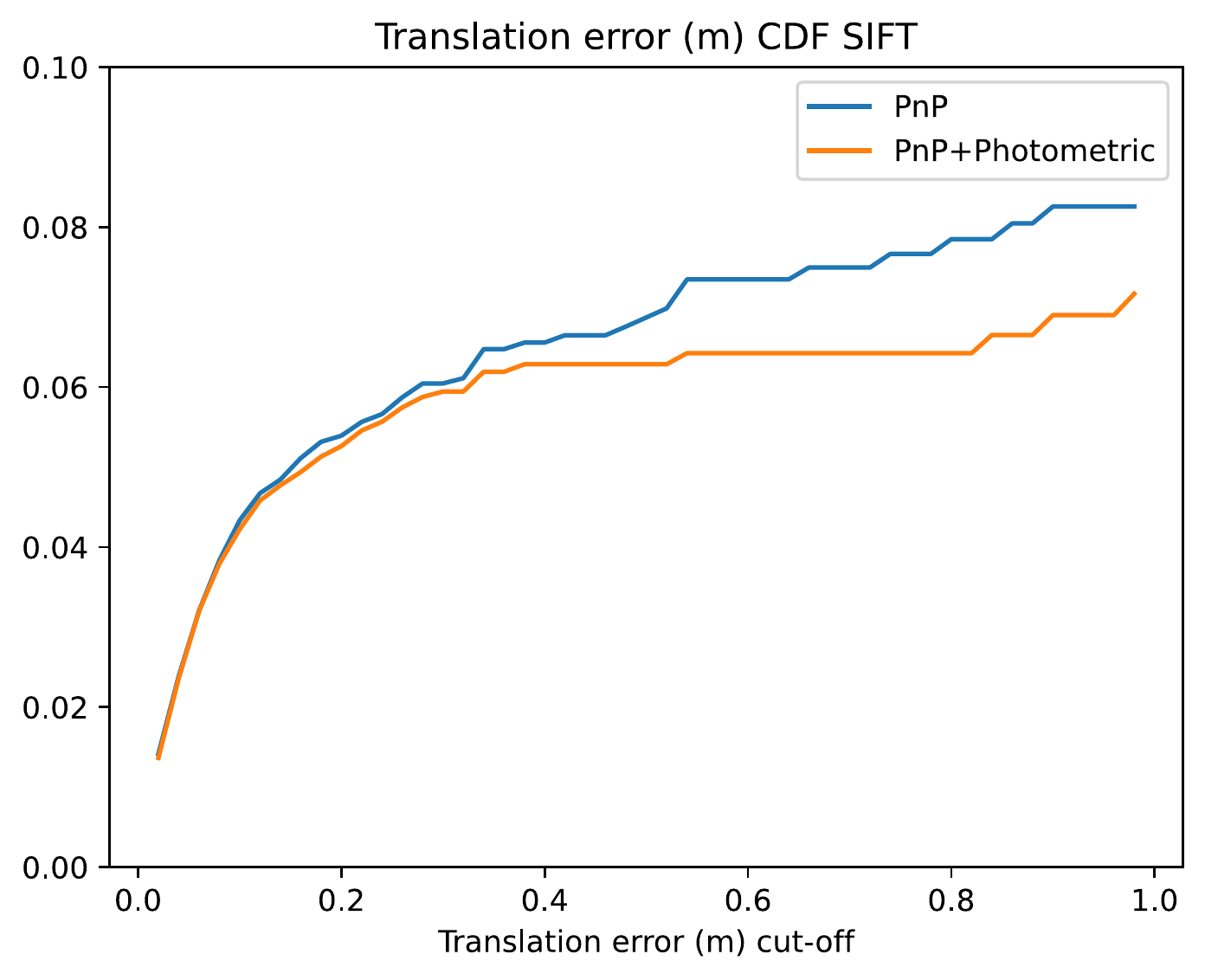} \\
  \includegraphics[width=0.3\textwidth]{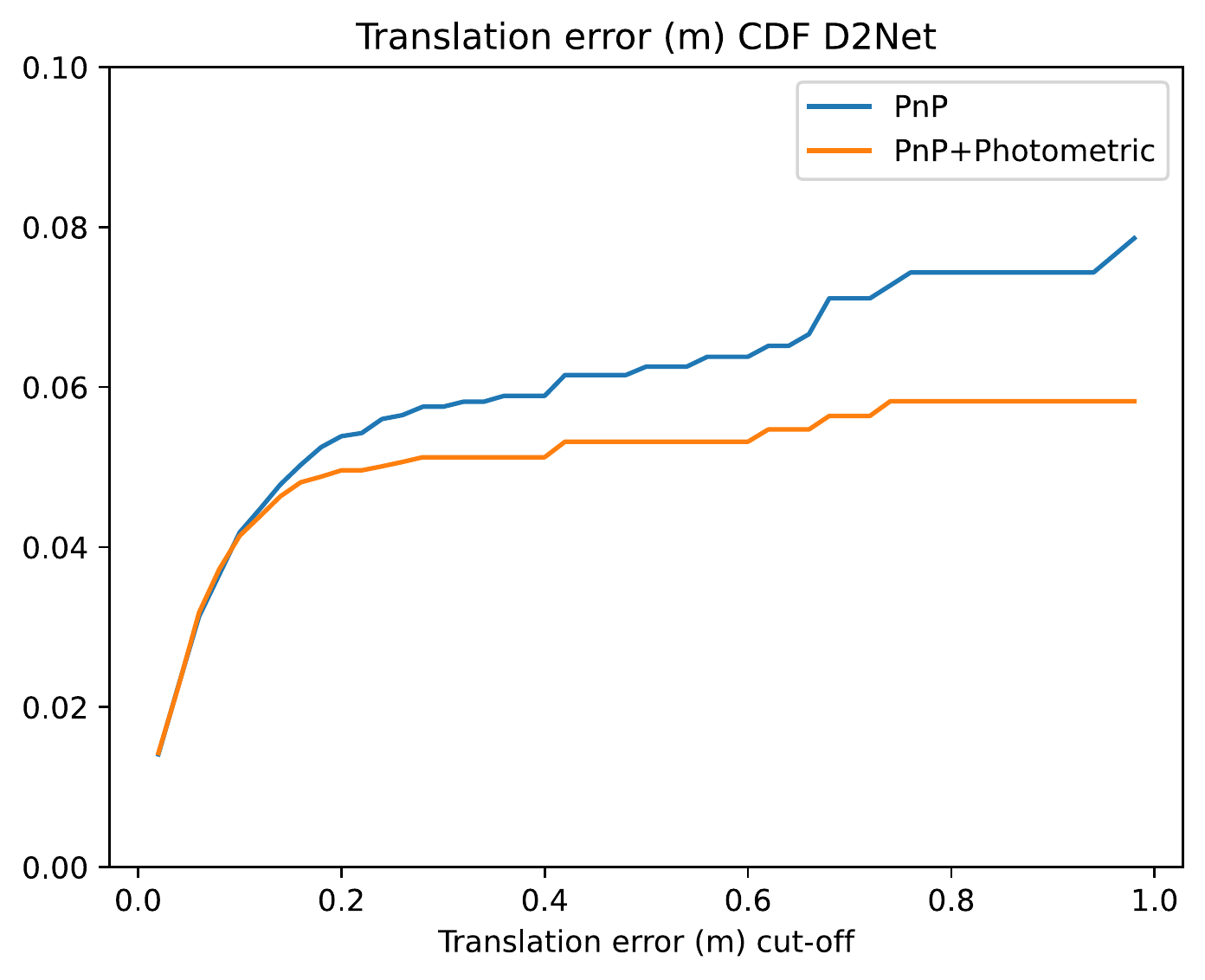} \\
  \includegraphics[width=0.3\textwidth]{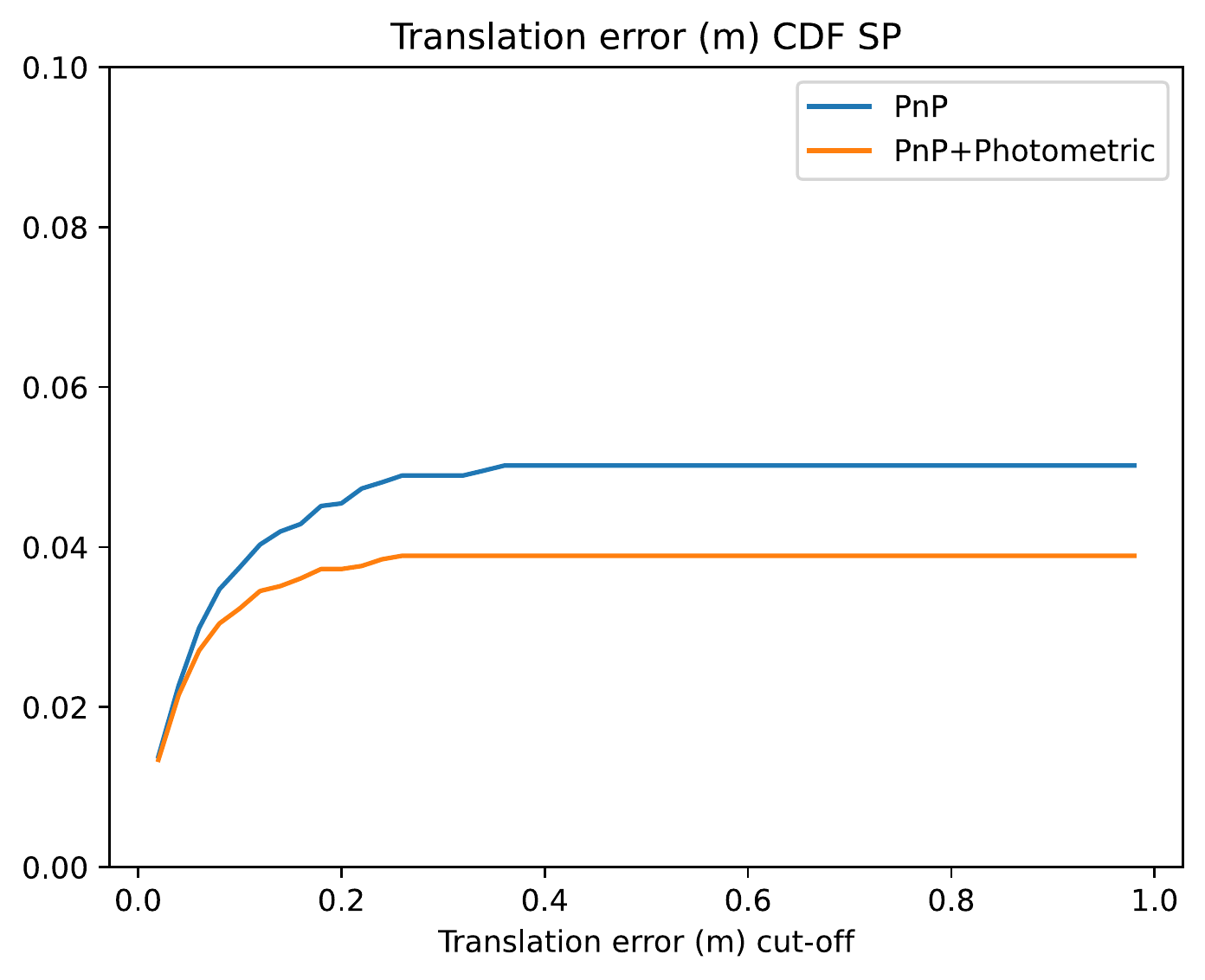} \\
  \end{tabular}
\caption{\textbf{Cumulative Error Plots for Various Local Features}. Cumulative error plots for (top) SIFT (middle) D2Net and (bottom) SuperPoint are shown. The blue line shows vanilla PnP, and the orange line shows the joint optimization of PnP + Photometric error, which is enabled by Nerfels. As the difficulty of problem increases from left to right on the x-axis, the benefit is larger.
}
\label{qual_results}
\vspace{15.0em}
\end{figure}


\clearpage

\section{Implementation Details}
\textbf{Network Architecture} For constructing the neural rendering decoder, we follow a similar network architecture as done in \cite{mildenhall2020nerf} with the additional Nerfel code $d_{c}=64$, concatenated to the encodings of the queried points. The queried points and view directions are encoded using the positional encodings with $\mathcal{L}_{x} = 10, \mathcal{L}_{\phi} = 4$. The point encodings are fed to an $8$ layer FC network with $256$ hidden dimension with point encodings being skipped and concatenated to the features every $3$ layers. This is followed by concatenating the view positional encodings with a single FC layer and the outputs result from $\sigma$ head and an RGB ($\hat{r}$) head.

\textbf{Training Settings} The decoder $D_{NR}$ was trained over $800k$ iterations using Adam optimiser \cite{kingma2014adam} with a learning rate of $5e^{-4}$ and a decay factor of $0.1$ annealed through the training iterations. The size of an image in the decoder training phase is $80 \times 60$, and the number of rays batched in each iteration is set to $1024$. The number of coarse rays sampled during training is $64$ followed by $128$ fine rays using the stratified sampling approach. Within the volume rendering function a $0.2$ STD of noise is used for perturbing the radiance field. When collecting the Nerfel codes (Section 4), the radius of a Nerfel sphere $r_{s} = 0.3$ in metric measurement and given a set of keypoints $\mathcal{K}$ over an entire scene the threshold $t_{f}$ is adaptively chosen so that $N_{c} \approx 0.2|\mathcal{K}|$.

\textbf{Evaluation Settings} For both synthetic and real datasets, image pairs were chosen with overlap of $[0.4, 0.8]$ to simulate wide-baseline scenarios. The overlap value was computed by symmetrically taking the reference frame and re-projecting it's 3D points onto the query image and and vice versa. We check the percentage of the points that fall within the opposing frame and the average of both is the overlap value.

\onecolumn

\clearpage

\section{Nerfel Samples}

In the following we provide qualitative samples of Nerfels with frontal rotation poses. \\


\begin{figure*}[b]
\centering
\includegraphics[width=1\textwidth]{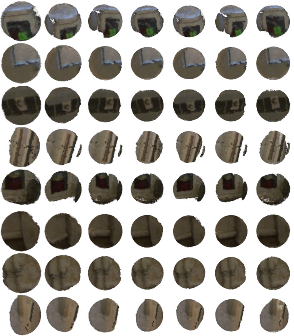}
\end{figure*}

\begin{figure*}
\centering
\includegraphics[width=1\textwidth]{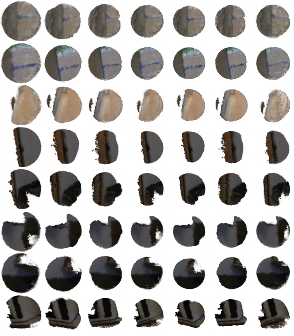}
\end{figure*}







\clearpage

\section{Nerf Full-Scene Rendering Samples}

Here we provide $5$ different locations in $4$ full-scenes that are implicitly represented using a Nerf network. For each location the camera was perturbed to show the surroundings. From the renderings it can be seen that attempting to model a full-scene using a Nerf with a similar capacity to that used for Nerfels results in low quality scene representation. This causes the pose-estimation optimisation converge to incorrect solutions.

\begin{figure}[h]
\centering
\includegraphics[width=1\textwidth]{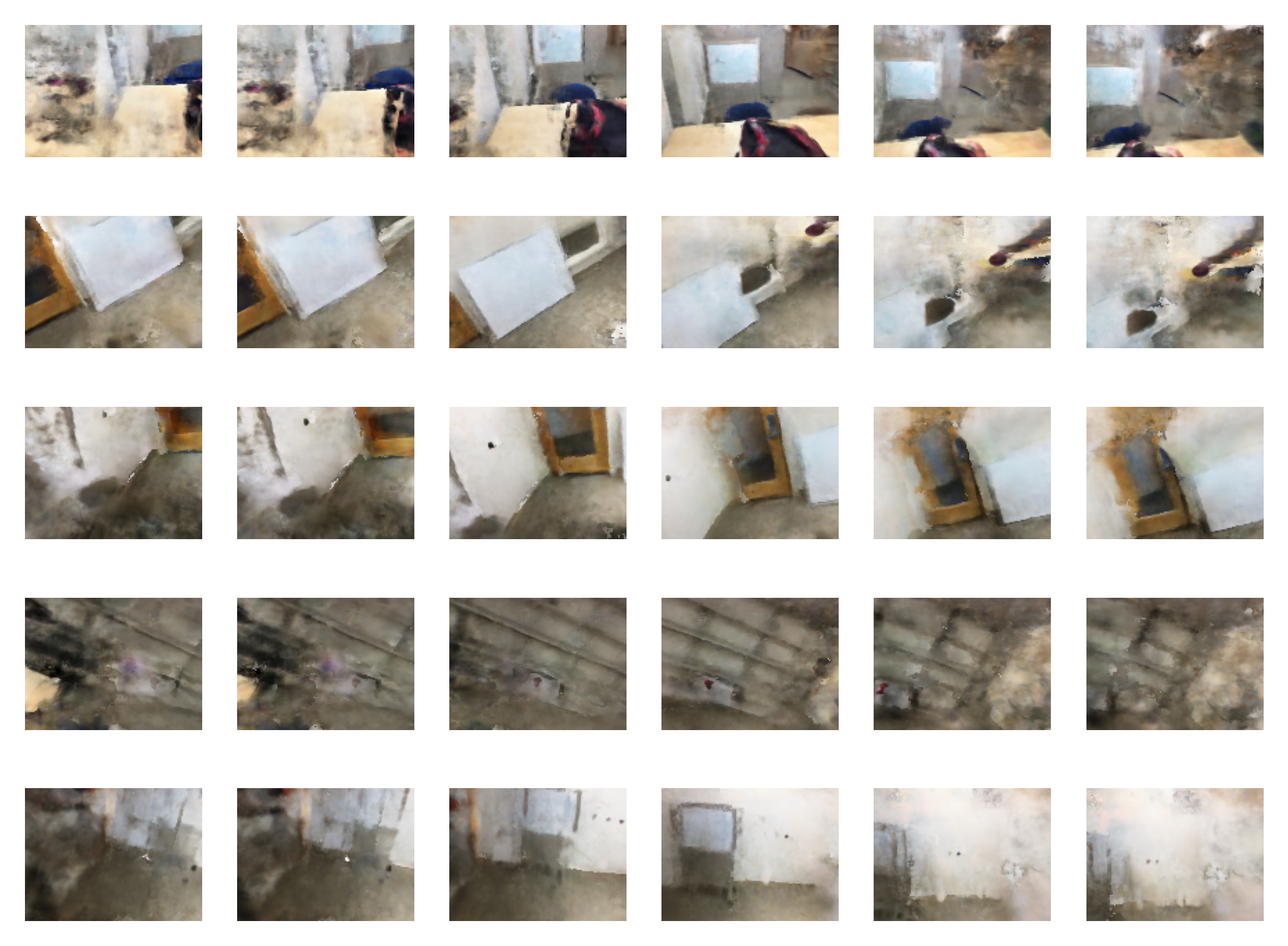}
\end{figure}

\begin{figure}[h]
\centering
\includegraphics[width=1\textwidth]{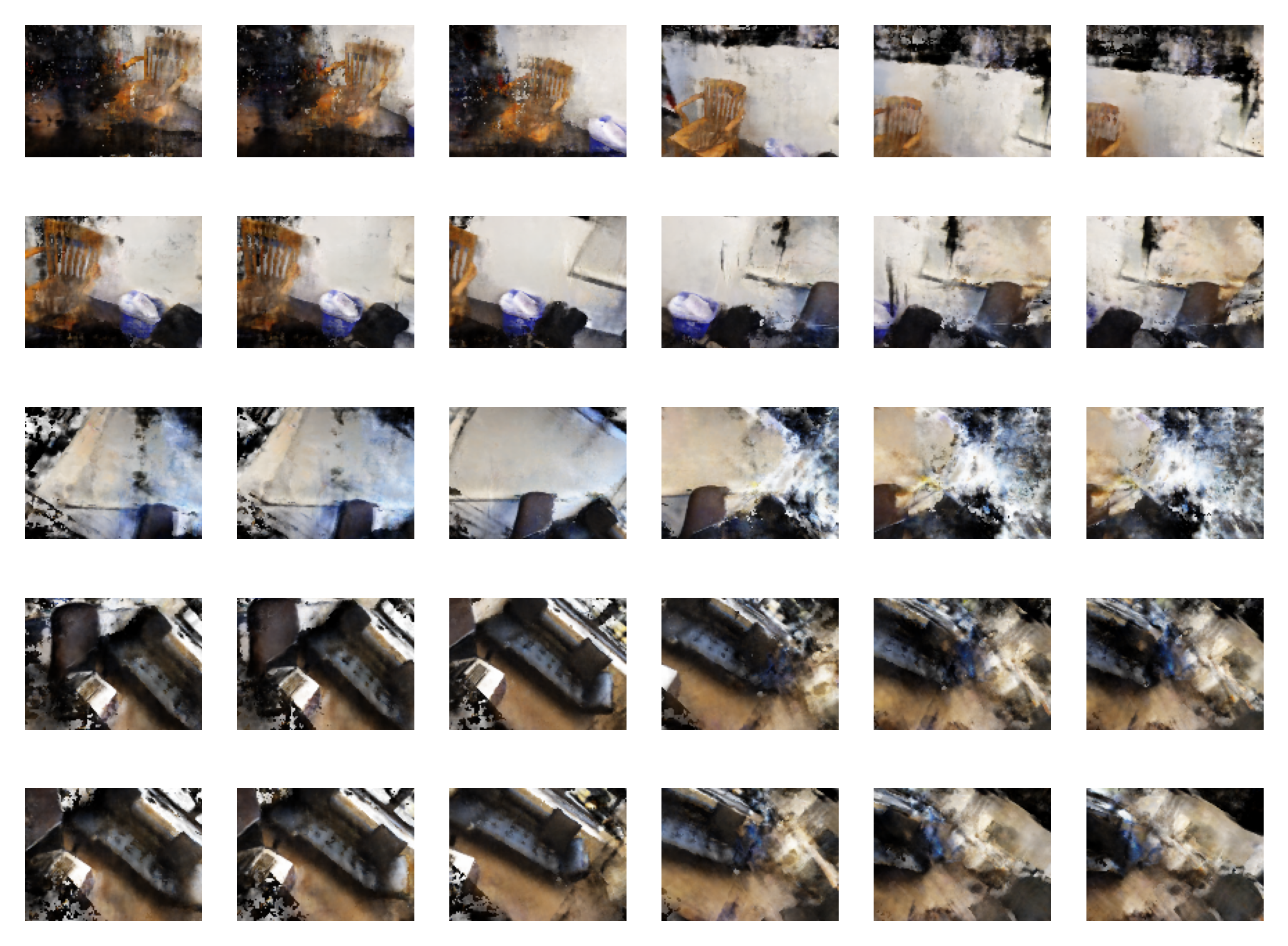}
\end{figure}



\twocolumn

\clearpage


\end{document}